\documentclass[journal]{IEEEtran}
\ifCLASSINFOpdf
   \usepackage[pdftex]{graphicx}
   \graphicspath{{./Figures/}}
\else
\fi

\usepackage[cmex10]{amsmath}
\usepackage{algorithmic}
\usepackage{algorithm}

\usepackage{widetext}
\usepackage{ulem}

\usepackage{draftwatermark}
\SetWatermarkText{IN PRESS}
\SetWatermarkScale{1}

\usepackage{makecell}

\usepackage{setspace}

\usepackage{array}
\newcolumntype{L}{>{\centering\arraybackslash}m{3cm}}

\usepackage{multicol}
\usepackage{widetext}

\usepackage[tight,footnotesize]{subfigure}

\usepackage{makeidx}  

\usepackage{multirow}
\usepackage{tabularx}

\usepackage{tikz}
\usetikzlibrary{shapes.geometric,arrows}

\tikzstyle{startstop} = [rectangle, rounded corners, minimum width=3cm, minimum height=0.25cm, text centered, draw=black, fill=red!30]
\tikzstyle{io} = [trapezium, trapezium left angle=70, trapezium right angle=110, minimum width=3cm, minimum height=0.5cm, text centered, draw=black, fill=blue!30]
\tikzstyle{process} = [rectangle, minimum width=3cm, minimum height=0.25cm, text centered, draw=black, fill=orange!30]
\tikzstyle{decision} = [diamond, minimum width=0.25cm, minimum height=0.25cm, text width=1.5cm, text centered,  draw=black, fill=green!30] 
\tikzstyle{arrow} = [thick,->,>=stealth] 

\begin{document}

\title{Design of OFDM radar pulses using genetic algorithm based techniques}

\author{Gabriel~Lellouch,~\IEEEmembership{}
        Amit~Kumar~Mishra,~\IEEEmembership{}
        and~Michael~Inggs,~\IEEEmembership{}

\thanks{G. Lellouch is with the Department
of Electrical Engineering, University of Cape Town, Cape Town, South-Africa.}
\thanks{A.K. Mishra and M. Inggs are with the Department
of Electrical Engineering, University of Cape Town, Cape Town, South-Africa.}
}


\maketitle

\begin{abstract}
The merit of evolutionary algorithms (EA) to solve convex optimization problems is widely acknowledged. In this paper, a genetic algorithm (GA) optimization based waveform design framework is used to improve the features of radar pulses relying on the orthogonal frequency division multiplexing (OFDM) structure. 
Our optimization techniques focus on finding optimal phase code sequences for the OFDM signal. Several optimality criteria are used since we consider two different radar processing solutions which call either for single or multiple-objective optimizations. 
When minimization of the so-called peak-to-mean envelope power ratio (PMEPR) single-objective is tackled, we compare our findings with existing methods and emphasize on the merit of our approach. 
In the scope of the two-objective optimization, we first address PMEPR and peak-to-sidelobe level ratio (PSLR) and show that our approach based on the non-dominated sorting genetic algorithm-II (NSGA-II) provides design solutions with noticeable improvements as opposed to random sets of phase codes. We then look at another case of interest where the objective functions are two measures of the sidelobe level, namely PSLR and the integrated-sidelobe level ratio (ISLR) and propose to modify the NSGA-II to include a constrain on the PMEPR instead. In the last part, we illustrate via a case study how our encoding solution makes it possible to minimize the single objective PMEPR while enabling a target detection enhancement strategy, when the SNR metric would be chosen for the detection framework.
\end{abstract}


\section{Introduction}

With the advent of powerful digital hardware, software defined radio and radar have become an active area of research and development \cite{Langman13}, \cite{Garmatyuk11}. This in turn has given rise to many new research directions in the radar community, which were previously not comprehensible. One such direction is the recently investigated OFDM radar \cite{Lellouch14c} which uses OFDM waveforms instead of the classic linear frequency modulated (LFM) waveforms. 

In communications, OFDM is a special form of multicarrier modulation (MCM), where a single data stream is transmitted over a number of lower rate subcarriers. Alphabets such as binary phase shift keying (BPSK), quadrature phase shift keying (QPSK), etc., are commonly used to code the information. The resulting complex symbols, also called phase codes (-1,+1 in BPSK) are modulating the subcarriers \cite{Hara03}. When the signal gets to the receiver, a demodulation stage retrieves the transmitted phase codes and eventually the binary message. In radar, the priority is to detect the presence of targets and possibly estimate some of their features through the following measurable quantities: range, Doppler, azimuth and elevation. It is thus of utmost interest for the radar designer to understand the possibilities offered by the OFDM structure and optimize it to suit its needs.

It can be quickly verified that non-coded OFDM pulses (all phase codes are 1 or at least equal) will not be suitable to radar systems that operate with the conventional matched filter processing, since they will give rise to high sidelobes. Another drawback of non-coded OFDM pulses lies in the behaviour of the time domain signal. Strong variations are detrimental since they can cause distortions in power amplifiers. Therefore, the OFDM pulse needs to be tailored before it becomes a suitable radar waveform.

In this work, we argue that 
some simple and easy-to-implement memetic computing techniques can be used to design OFDM radar pulses. Note that the investigation of new evolutionary techniques is out of the scope. We look at the problem of reducing the peak-to-mean envelope power ratio~(PMEPR) solely and jointly with the problem of reducing the autocorrelation sidelobes.  These aspects respond to two processing mechanisms specific to OFDM radar that were presented in~\cite{Lellouch14c}. One of these mechanisms, so called frequency domain processing, mostly applies to tracking scenarios. When dealing with the sidelobes' reduction (in regard of the other mechanism), because our approach is restricted to the range dimension it remains applicable to tracking scenarios where the Doppler effect can be accounted for. In this context, one would not consider the autocorrelation function but rather, the correlation of the pulse with the same pulse including Doppler. Although a number of techniques have addressed these objectives separately and jointly~\cite{Mozeson03,Levanon04} they also give rise to several limitations. For example, the Newman phasing technique~\cite{Newman65} gives a PMEPR below 2 for the single OFDM symbol case, and this for most numbers of subcarriers $N$~\cite[chap.~11]{Levanon04}, but it degrades when the subcarriers are not contiguous as we shall see. Another drawback is the impracticality to reuse existing communication schemes that, for instance would rely on predefined alphabets such as BPSK, QPSK, etc.. The third limitation regards the lack of control over the spectrum. This is detrimental in the context of matched illumination, which is a strong incentive to use OFDM radar~\cite{Genderen09}. Hence, this paper aims to feature emerging evolutionary algorithms (EA) in response to the above challenges and includes multiple-objective optimization (MOO) considerations, which have not been researched before.

When addressing the single-objective optimization (e.g. PMEPR), we implement the genetic algorithm (GA) in its binary~\cite{Whitley94} or continuous form~\cite{Haupt04}. The binary form is well suited when working with discrete sets of phase codes (i.e. constellations). In the case of multiple-objective optimizations, we use the well known non-dominated sorting genetic algorithm II (NSGA-II)\cite{Deb02}. It has proven to work much faster than the earlier version~\cite{Deb01}, NSGA, while providing diversity in the solutions. Other radar related studies have implemented the same algorithm \cite{Sahoo09,Sahoo13,Sen11}.   

In the context of OFDM signalling for radar problems, our contribution lies in the use of genetic algorithm based methods, where the parameters are the real arguments of the complex phase codes of the OFDM signal, to control the OFDM pulse in terms of its time domain variations and its autocorrelation sidelobes. To the authors' best knowledge, these considerations are novel and contribute to the field of OFDM radar whose state-of-the art as reviewed in~\cite{Genderen09,Genderen10} is still applicable. To compare with more recent works, it must be noted that in~\cite{Sen11}, Sen et al. address a detection problem in the condition of multipath, in which MOO-GA based techniques are employed to optimally design the spectral parameters of the OFDM waveform for the next coherent processing interval. There, the objective functions are driven by optimal detection considerations and therefore are distinct from those considered in this paper. 

The rest of the paper is organized as follows. In Section~\ref{sec:sect2} we aim at dimensioning the OFDM pulse that we will then optimize, in particular, we assess realistic values for the number of subcarriers. For this, we briefly review the processing solutions proposed in OFDM radar and connect the OFDM pulse parameters to external parameters, after formulating constraints driven by the processing and the scenario. In Section~\ref{sec:sect3} we review the objective functions at stake in the specific case of discrete signals, recall their relevance with respect to the processing technique selected and discuss the variables involved in the optimization. In Section~\ref{sec:sect4}, we present our optimization techniques, starting with a review of state-of-the-art methods followed by a discussion on the incentives to employ genetic algorithm based procedures that builds upon the limitation of the existing techniques. We then review swiftly our encoding strategy. In Section~\ref{sec:sect5}, we present our numerical results for the various cases of single objective and multiple objectives and establish the merit of the techniques by comparing with prior studies, when possible. 
In Section~\ref{sec:sect6}, we address the specific case of PMEPR optimization as part of a matched illumination procedure. We demonstrate that, on the one hand, the target detection enhancement problem and, on the other hand, the PMEPR optimization problem, can be solved if we reserve the real weights to address the first problem and use the complex phase codes for the other. We give our conclusions in Section~\ref{sec:sect7}.

\section{Dimensioning the OFDM pulse}\label{sec:sect2}   
In this section, we present the successive steps that form our design strategy for fixing some of the OFDM pulse parameters. We show how they are inferred from, on the one hand, the type of processing and on the other hand, the scenario. 
This step is particularly relevant to reveal some specificities of pulsed OFDM radar, as opposed to continuous OFDM radar. The results from this section are used later on to generate pulses with appropriate dimensions. 

\subsection{Processing related constraints}\label{sec:sect2a}
In the scope of a pulsed OFDM radar waveform, we proposed in \cite{Lellouch14c} two processing alternatives. The first alternative is based on the combination of matched filtering and Doppler processing while the second alternative transforms the received signal in the frequency domain in the same way OFDM communication systems operate. After a demodulation stage that suppresses the phase codes, two orthogonal discrete Fourier transforms (DFT) are applied to form a range Doppler image. The key characteristics of both processing are recalled in Table~\ref{tab:processing_charact}. In the rest of the paper we refer to the former as the time domain processing while we define the latter as our frequency domain processing. 

\begin{table}[h]
\caption{Characteristics of our OFDM processing alternatives} 
\centering 
\setlength{\extrarowheight}{1.5pt}
\begin{tabular}{c|p{3.7cm}|p{3.1cm}}
 & \textbf{Time domain processing} & \textbf{Frequency domain processing} \\\hline
 \multirow{2}{*}{\textbf{Pros}}&Immune to intersymbol interference & Range and Doppler sidelobes are phase codes independent\\
    &Doppler sidelobes are phase codes independent \\
    \hline
\textbf{Cons}&Range sidelobes are phase codes dependent &Subject to intersymbol interference\\ 
\end{tabular}
\label{tab:processing_charact}
\end{table}

The frequency domain processing is subject to inter-symbol interference (ISI). 
This issue has been of utmost interest in the early years of OFDM signalling for communication, to cope with the multipath effect. To that end the concept of cyclic prefix has been introduced \cite{Hara03}. In pulsed OFDM radar, cyclic prefix is no longer a relevant feature, as it is in its continuous counterpart \cite{Tigrek12a}, hence in this discussion we propose to mitigate the risk of ISI by matching 
the size of the range cell according to the target extent. This supposes prior information over the target dimension, thus we recommend using this processing for tracking. Since the time domain processing does not come up with a severe design constraint we choose to use the constraint resulting from the frequency domain processing as the main guideline to dimension the OFDM pulse.

\subsubsection{Sampling frequency}\label{sec:sect2a1}
Assuming that we work with the complex baseband OFDM signal in the receiver, with bandwidth $B$, 
the Nyquist theorem states that the sampling frequency can be taken as low as $f_s=B$ and the time cell size is thus inversely proportional to the bandwidth, $t_s=1/B$. The size of the range cell is then given by $c/2B$, where $c$ is the speed of light. 

\subsubsection{Bandwidth}\label{sec:sect2a2}
\begin{figure}[!ht]
\centering
\includegraphics[scale=0.30]{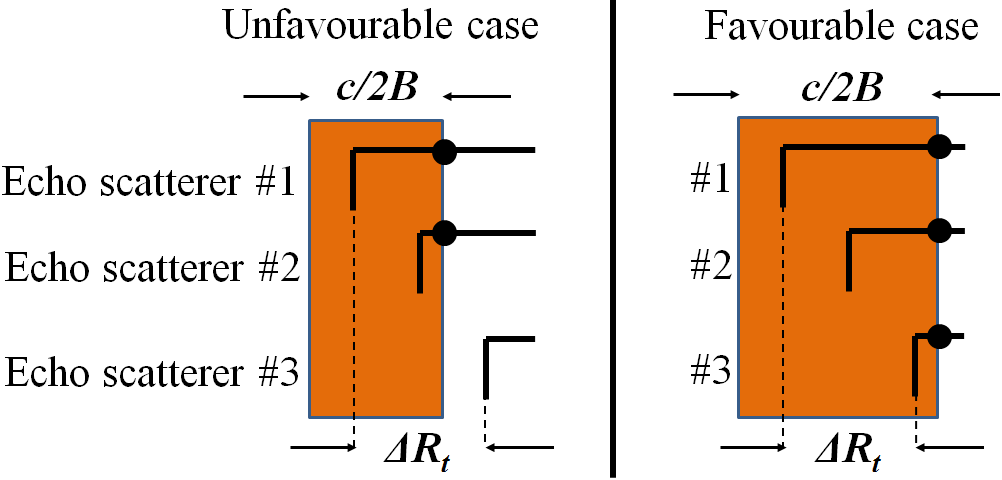} 
\caption{Constraint applicable to the size of the range cell in order to mitigate the risk of ISI in the case of the frequency domain processing. Echoes received from the different scatterers that compose the target shall fall within the same range cell.} 
\label{fig:ISI_constraint_graph}
\end{figure}

To satisfy the above constraint on the size of the range cell 
we shall adjust the bandwidth to comply with: $c/2B\geq \Delta R_t$ where $\Delta R_t$ is the target range extent. Practically we can add a margin to account for the target radial velocity and the uncertainty on the target extent and position. Figure~\ref{fig:ISI_constraint_graph} illustrates this constraint, which is itself inferred by the loss of orthogonality of the subcarriers caused by ISI~\cite{Lellouch14e}. Not only the received echoes shall remain in the same time cell as a result of the first pulse but also throughout the coherent processing interval. In the end the bandwidth can be obtained from:
\begin{equation} \label{eq:Bandwidth}  
B = \frac{c}{2(\Delta R_t+\alpha)}
\end{equation}
where $\alpha$ is the margin in range.

Therefore, if there is no need for high range resolution (assuming that the target is known) we suggest to dimension the bandwidth based on these ISI considerations. 

\subsection{Scenarios related constraints}\label{sec:section2b}
Other parameters need to be fixed, viz., the pulse length and the number of subcarriers.

\subsubsection{Pulse length}\label{sec:section2b1}
To dimension the pulse length, we relate it to 
the eclipsed zone, the window that lies between the radar and the minimum detection range like it is commonly done for pulsed radar. 
If, say, we expect targets from $R_{\text{min}}=1.5$~km, an upper bound for the pulse length is found to be \cite{Richards10}, $t_p=2 R_{\text{min}}/c=10$~$\mu$s. 

\subsubsection{Number of subcarriers}\label{sec:section2b2}
The orthogonality property is another example of the unique OFDM structure. It states that the symbol or bit duration $t_b$ is inversely proportional to the subcarrier spacing $\Delta f$, $t_b=1/\Delta f$. In the extreme case where the pulse is composed of one symbol, the maximum number of subcarriers $N_{\text{max}}$ used in the pulse can be derived from: 
\begin{equation} \label{eq:Nmax}  
N_{\text{max}} = \frac{2BR_{\text{min}}}{c},
\end{equation}
where we used the previous result for the pulse length upper bound. For the same pulse bandwidth and pulse duration a smaller number of subcarriers can be used if we construct the pulse from several symbols. For example, we can decide to use 250 subcarriers with 4 symbols in the pulse to maintain the same duration. The subcarrier spacing is then increased from 100~kHz to 400~kHz. Information related to the target maximum speed shall also be considered to ensure negligible distortion effects in either processing \cite{Lellouch14b,Lellouch14c}.  

To put the above analysis into perspective, we suggest two objects of different size, on the one hand a walker and on the other hand, a truck. Table~\ref{tab:scenar_charact} summarizes the values attributed to the bandwidth and the maximum number of subcarriers for both cases.

\begin{table}[!ht]
\caption{Scenarii characteristics for the waveform design} 
\centering 
\begin{tabular}{c | c | c} 
 & Case 1 & Case 2 \\ [0.5ex]
 & (walker) & (truck) \\ [0.5ex] 
\hline  
Range extent (m)& 2 & 10 \\ 
Margin (m) & 1 & 5 \\
\hline 
\textbf{Bandwidth (MHz)}& \textbf{50} & \textbf{10} \\ 
\textbf{Maximum number of subcarriers}& \textbf{500} & \textbf{100} \\ 
\end{tabular}
\label{tab:scenar_charact} 
\end{table}

\section{Optimizing the pulse for radar}\label{sec:sect3}

Having dimensioned our OFDM pulses we are now ready to concentrate our analysis on the optimization of those OFDM pulses for radar. Firstly, we need to define our objective functions and secondly, we need to identify the OFDM parameters or variables upon which we can run our optimization procedure.  

\subsection{Objective functions}\label{sec:sect3a}
\begin{figure}[!ht]
\centering
\mbox{\subfigure[LFM pulse]{\includegraphics[width=1.75in]{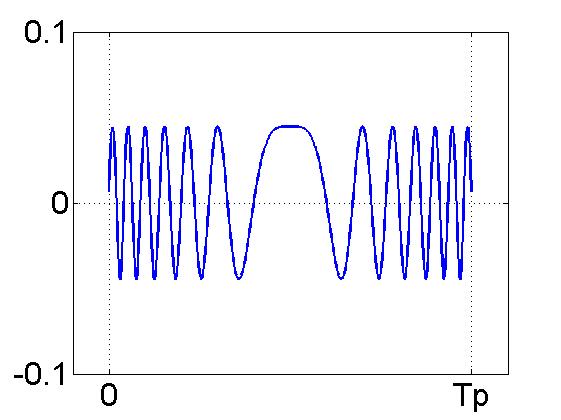}\label{fig:LFM_pulse}} \subfigure[OFDM pulse]{\includegraphics[width=1.75in]{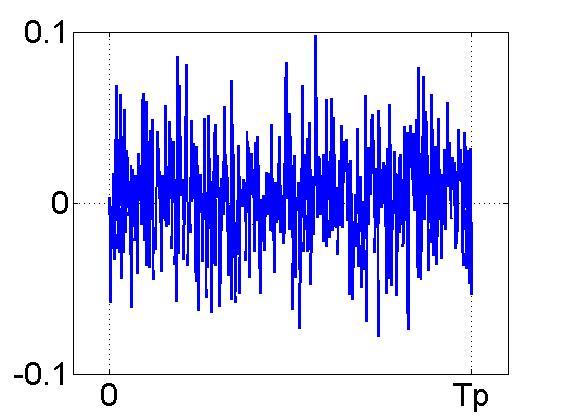}\label{fig:OFDM_pulse}} }
\caption{Real part of a) a LFM pulse versus b) an OFDM pulse with random starting phases. Both pulses have been normalized to have unit energy.} \label{fig:LFM_vs_OFDM_pulse}
\end{figure}

An OFDM symbol is built as a sum of weighted complex sinusoids, where every sinusoid has a given starting phase. Because the multiple frequencies (or subcarriers) are transmitted simultaneously the full bandwidth is covered at every instant in the OFDM configuration~\cite{Levanon02a}. Fig.~\ref{fig:LFM_vs_OFDM_pulse} depicts an OFDM pulse and compares it with the typical LFM pulse. 

When the OFDM pulse is composed of several symbols it can be expressed as:
\begin{equation} \label{eq:OFDMbaseband}  
x(t) = A\sum_{n=1}^{N} w_n\left(\sum_{k=1}^{K}  a_{n,k} r_k(t)\right) \exp(j2\pi n\Delta f t).
\end{equation}
$N$ is the total number of subcarriers in the pulse, $w_n$ denotes the weight applied on subcarrier $n$ and $a_{n,k}$ is the phase code attributed to subcarrier $n$ in symbol $k$. $K$ is the total number of symbols in the pulse. The function $r_k(t)$ refers to the rectangular window for every symbol: 
\[ r_k(t) = \left\{
  \begin{array}{l l}
    1 & \quad kt_b \leq t< (k+1)t_b\\
    0 & \quad \text{elsewhere}
  \end{array} \right.\]

Like in \cite{Lellouch14e}, the normalization factor $A$ is given by:
\begin{equation} \label{eq:normalization_factor}  
A = \frac{1}{\sqrt{\sum_{k=0}^{K-1}\sum_{n=0}^{N-1}w_n^2|a_{n,k}|^2t_b}}.
\end{equation}

When using OFDM for radar two important aspects must be considered.

\subsubsection{Peak-to-mean envelope power ratio}\label{sec:sect3a1} 
This aspect is specific to OFDM signals. It 
characterizes the variations in time of the signal envelope as observed in Fig.~\ref{fig:OFDM_pulse}. Because they are generally used in saturation to maximize the transmitted power, devices like power amplifiers can distort the signal and be detrimental to the system performance. 
While the peak-to-average power ratio (PAPR), defined as the ratio of the peak power to the average power of the real-valued multicarrier signal, is usually of interest, we consider the peak-to-mean envelope power ratio (PMEPR) because we work with complex baseband signals. By definition, the PAPR will result in smaller values than the PMEPR \cite{Levanon04} whose expression is given by \cite{Jones94}:
\begin{equation} \label{eq:PMEPR} 
PMEPR = \frac{\underset{n}{\operatorname{max}} |x[n]|^2}{\frac{1}{N}\sum|x[n]|^2},  
\end{equation}

\subsubsection{Sidelobe level}\label{sec:sect3a2}
Typically, with the time domain processing where matched filtering is applied in range, it is important to keep the sidelobes of the compressed pulse as low as possible. 
Hence, our next objective functions regard the behaviour of 
the autocorrelation function, whose expression $R(\tau)$ is given by:
\begin{equation} \label{eq:Xcorr_cont} 
R(\tau) = \int_{-\infty}^{\infty}x(t)x^*(t-\tau)dt.  
\end{equation} 
and in its discrete form, $R[m]$:
\begin{equation} \label{eq:Xcorr_disc} 
R[m] = \sum_{p=0}^{NK-1}x[p]x^*[p-m],  
\end{equation} 
where $m$ takes integer values between $-NK+1$ and $NK-1$. $x[p]$ represent the discrete values of the OFDM pulse taken at the discrete instants $pt_b/N$, where $p$ takes integer values from 0 to $NK-1$. 
If the pulse is composed of only one symbol, then $p$ takes values from 0 to $N-1$, just like $n$ the subcarrier index. In the end, $x[p]=x(pt_b/N)$. In Eq.~\ref{eq:Xcorr_disc}, we assume that $x[p]=0$ for all forbidden values of $p$, that is $p<0$ and $p>NK-1$.

To cope with practical applications we commonly distinguish two objective functions. The first function is the PSLR. It returns the ratio between the highest sidelobe and the peak.
\begin{equation} \label{eq:PSLR} 
PSLR = \frac{\underset{m}{\operatorname{max}}  |R[m]|}{|R[0]|}, m\neq 0  
\end{equation} 
The second function is the integrated-sidelobe level ratio (ISLR). It returns the ratio between the cumulation of the sidelobes and the peak. 
\begin{equation} \label{eq:ISLR} 
ISLR = \frac{\underset{m}\sum |R[m]|}{|R[0]|}, m\neq 0 
\end{equation}
The weight attributed to both figure-of-merits depends on the application as well as the environment. For example, if the radar operates in the presence of distributed clutter, it will be important to work with low ISLR in order to keep the weak targets visible. In that case, high ISLR can be interpreted as an increase of the noise floor. Conversely if the application requires detection of targets in the presence of strong discrete clutter, the PSLR is more critical and must be kept low to prevent from deceptively considering one sidelobe as another small target. In Section~\ref{sec:sect5}, we present our results using the conventions $PSLR_{\text{dB}}$~=~$20\log(PSLR)$ and $ISLR_{\text{dB}}$~=~$20\log(ISLR)$. Remark that the above objective functions are for the zero Doppler case. In fact, would the design concern a tracking scenario, the Doppler effect could be accounted for in the model of the received signal such that the objective function would no longer be inferred from the autocorrelation function but simply the correlation of the pulse with the pulse including Doppler. In the case where there is no a priori information concerning Doppler, the optimization shall be done over the range Doppler plane that includes all possible Doppler values. In~\cite{Lellouch14c} we showed that a 1~dB compression loss on the main peak would arises when $f_d$$>$$\Delta f/K$, hence if the subcarrier spacing is large enough (few MHz) the OFDM pulse is Doppler tolerant. However, this does not conclude on the sidelobe level over the range Doppler domain. Considerations regarding the optimization of the range Doppler sidelobes are outside the scope of this paper.

\subsubsection{Need for oversampling}\label{sec:sect3a3}
\begin{figure}[!ht]
\centering
\mbox{\subfigure[Time domain OFDM pulse]{\includegraphics[width=1.8in]{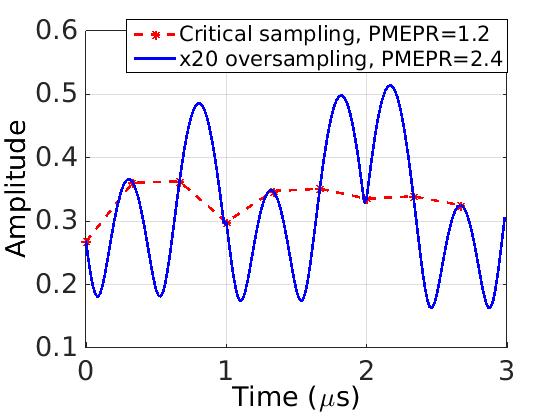}\label{fig:PMEPR_oversamp_demo}}
\subfigure[Autocorrelation output]{\includegraphics[width=1.8in]{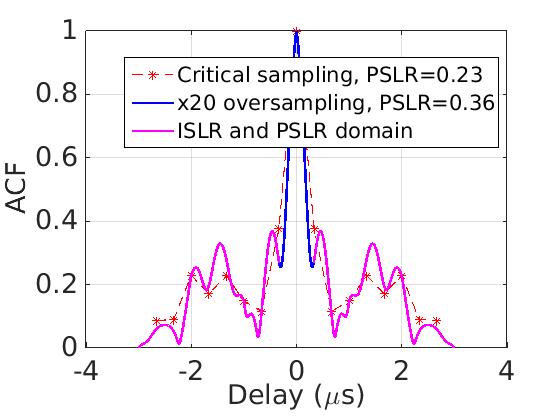}\label{fig:ACF_oversamp_demo}} }
\caption{a) Amplitude versus time of an OFDM pulse and b) its autocorrelation function versus range when $N$=3 and $K$=3 in the case of critical sampling $f_s$=$B$ and oversampling $f_s$=20$B$. The values obtained for the PMEPR and the PSLR are given.} \label{fig:ACF_PMEPR_demo}
\end{figure}

In Eq.~\ref{eq:PMEPR}, Eq.~\ref{eq:PSLR} and Eq.~\ref{eq:ISLR}, we have assumed to work at the critical sampling rate $f_s=B$ such that the sampling period is $t_s=t_b/N$, as a result of the relationships that govern the OFDM structure. In Fig.~\ref{fig:ACF_PMEPR_demo} we emphasize on the need for oversampling. Because of the quick temporal variations of the OFDM signal, the PMEPR of the critically sampled signal will differ from the PMEPR of the continuous signal. An important question is how large the oversampling factor should be in order for the approximation to be accurate. In \cite{Tellambura01}, the difference between the continuous time and discrete time PAPR is evaluated and the conclusion is that an oversampling factor of 4 is accurate for an OFDM signal with BPSK coding. 
In this work, we choose to oversample the OFDM signal with a factor of 20 before evaluating our objective functions. In Fig.~\ref{fig:ACF_oversamp_demo}, we show how both PSLR and ISLR calculations exclude the values around the main peak. The total peak extent is equal to twice the Rayleigh resolution, which in time is $2/B$ \cite{Lellouch14a}.

\subsection{Single or multiple objectives}\label{sec:sect3b}  

In light of the above, we see that in the process of designing an OFDM pulse, extra care shall be employed to maintain tolerable PMEPR as well as reasonable sidelobe levels. 
However, if we rely on our frequency domain processing, we have recalled in Table~\ref{tab:processing_charact} that both the range and Doppler sidelobes~\cite{Lellouch14b} were phase code independent. In this case, the pulse design can focus on the PMEPR only. In a nutshell, with pulsed OFDM radar, we suggest two distinct strategies for the design of the pulses, viz., a single objective PMEPR optimization for the frequency domain processing and a multi-objective PMEPR and autocorrelation sidelobes' level optimization for the time domain processing. 

\subsection{Variables for the optimization}\label{sec:sect3c}
The parameters that remain after dimensioning the OFDM pulse are, as seen in Eq.\ref{eq:OFDMbaseband}, the weights $w_n$ and the phase codes $a_{n,k}$. Together, they form a vector of complex numbers $w_n\exp(j\phi_{n,k})$, where $a_{n,k}$=$\exp(j\phi_{n,k})$ and $\phi_{n,k}$ are real numbers between 0 and 2$\pi$. As mentioned earlier, the weights can also be regarded as spectral coefficients and as a result of the scattering mechanism observed in radar, their value may influence the detection process. Hence, it may be favourable to disregard them as variables for the optimization of the above objective functions. 

\section{Optimization techniques}\label{sec:sect4}
In this section, we first revisit state-of-the-art techniques that addressed the objective functions presented earlier. Then, we discuss their limitations and motivate the use of genetic algorithm based methods. 

\subsection{State-of-the-art methods}\label{sec:sect4a}
The problems of minimizing the PMEPR and the PSLR have been extensively researched since the emergence of multicarrier radar signals. 
Two strategies have stood out~\cite[chap.~11]{Levanon04}. The first approach is to consider identical sequences (IS), such that all subcarriers are assigned the same phase code. Optimizing the PMEPR of the pulse results in optimizing the PMEPR of a single symbol. In this context, Newman~\cite{Newman65}, Schroeder~\cite{Schroeder70} and Narahashi~\cite{Narahashi94} have suggested different phasing methods, where the weights $w_n$ are complex values with quadratic dependence on $n$. PMEPR as low as 2 (3~dB) can be obtained for any relevant value of $N$ (up to 65,000). Sidelobe levels of the autocorrelation function (ACF) can then be adjusted by introducing a linear carrier phase term that does not impinge the PMEPR but can change the ACF sidelobe pattern~\cite{Levanon04}. The second design strategy is based on modulating all $N$ subcarriers with consecutive ordered cyclic shifts (COCS) of an ideal chirplike sequence (CLS) of length $K$, assuming that $K$ symbols compose the pulse. For example, an OFDM pulse based on COCS of P4 codes can give PMEPR below 2 and PSLR below -15~dB for a large range of $N$ (between 0 and 70). Besides, note that a handful of methods aiming to reduce the PMEPR of OFDM symbols for communication have been developed~\cite{Han05}, however they refer to the transmission of information and hence fall outside the scope of this discussion.

\subsection{Motivation for genetic algorithm optimization methods}\label{sec:sect4b} 
Although the aforementioned techniques produce excellent results in terms of PMEPR alone or PMEPR and PSLR combined, they suffer from several limitations that may be detrimental in prospective situations of interest. Firstly, those techniques need contiguous subcarriers: from 1 to $N$. Hence, what if we want to use a pulse with a sparse spectrum (as a result of jamming, interference or simply frequency allocation)~\cite{Guo14}? The objective functions will be degraded undoubtedly. Secondly, they imply that existing communication schemes (such as BPSK, QPSK etc. constellations) cannot be used, which may prevent from building upon existing communication infrastructure to perform radar. Thirdly, the weight vector is fixed in the optimization mechanism of those objective functions. This prevents from shaping the spectral components of the signal to improve the detectability of an object whose radar cross section~(RCS), we know, resonates variably at different frequencies~\cite[chap.~7]{Richards10}, \cite{Sen10}. In this context, GA based methods offer favourable optimization strategies for they are able to address these aspects and are easy to implement. Other incentives include the possibility to work with modified objective functions, add new objective functions such as ISLR, propose a multitude of 'good' solutions and provide Pareto fronts of optimal solutions in a multi-objective sense.
Until now, just a few authors have considered the use of GA and MOO techniques for radar, however an increasing number of papers have recently established the potential of these optimization methods, in the context of improved detection under multipath conditions for OFDM radar~\cite{Sen11}, range ambiguity suppression in OFDM SAR~\cite{Riche12} and OFDM-STAP detection~\cite{Sen14}. Other optimization techniques that are receiving increasing attention within the broad engineering field are convex methods~\cite{Boyd04}. Although we do not claim that our problem is not convex or cannot be made convex, considerations regarding the use of convex optimization is beyond the scope of this paper. 

\subsection{Problem encoding}\label{sec:sect4c}
In our approach, the variables consist of the $NK$ arguments (real numbers) $\phi_{n,k}$. Note that the OFDM symbols result from applying an inverse discrete Fourier transform (IDFT) on each of the $N$ long vectors $[w_n\exp(j\phi_{n,k})]_{n=1\cdots N}$~\cite{Lellouch14c}. The OFDM pulse is then obtained by concatenating the symbols if $K>1$. The variables can either be encoded in the form of binary strings~\cite{Whitley94} or alternatively real values~\cite{Agrawal94,Haupt04}. In the case where we would employ a predefined constellation, it makes sense to use the first option since 'by nature', each phase code translates into a short string of few bits (e.g. two bits for QPSK, 3 bits for 8-PSK etc.). In the continuous case, the resolution depends on the system resolution, which in our case, allows us to represent one real number with 18 bits. The resolution in angle is then $\Delta\theta=2\pi/2^{18}\simeq0.024$~mrad. Note that we choose 18 to maximize the size of our domain of solutions and hence increase the chance to approach the best solution at the cost of increasing the search space. In practice, one may want to restrict it to a smaller number for the sake of reducing the computation complexity.

\section{Numerical results}\label{sec:sect5}
 
In this section we present our results for the various configurations, single and multiple-objective optimizations. We start with the case of the single objective PMEPR. Our intention is to show that the genetic algorithm is an elegant method to find optimal sets of phase codes, while challenging important limitations of state-of-the-art techniques. We then present our results for the case of multiple objectives and develop our discussion to introduce a PMEPR constrained multi-objective optimization where the original NSGA-II is slightly modified.
 
\subsection{Single objective: PMEPR}\label{sec:sect5a} 


In our simulations, we implemented a simple genetic algorithm (SGA)~\cite{Whitley94}. We considered a population of 12~chromosomes~\cite{Agrawal94} and included mutation (flipping one bit of the binary string that forms the chromosome, at random) at the rate of one mutation for 5 'offspring' chromosomes, selected at random every two generations. Despite the simplicity of our genetic algorithm, we are able to retrieve phase code sequences with optimal PMEPR properties.  

\begin{figure}[!ht]
\centering
\mbox{\subfigure[M-PSK (18 genes)]{\includegraphics[width=1.8in]{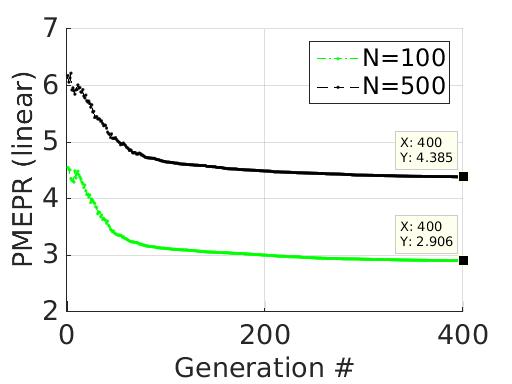}\label{fig:convergence_PMEPR_genes18}}
\subfigure[QPSK (2 genes)]{\includegraphics[width=1.8in]{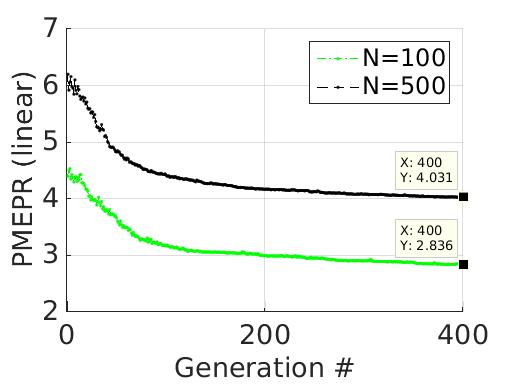}\label{fig:convergence_PMEPR_genes2}}}
\caption{Convergence of the SGA while encoding one phase with either a) 18 or b) 2 genes, for $N$=100 and $N$=500. A minimum PMEPR of 2.9 is achieved after 400 generations in average for both cases, when $N$=100. In fact, we identified a PMEPR as low as 2.6 and 2.7 in our populations after 400 generations, respectively for the M-PSK ($M$=$2^{18}$) and QPSK cases.} \label{fig:PMEPR_N_10_convergence}
\end{figure}

We have run a Monte-Carlo simulation with 20 runs and for each run at each generation, we found the minimum PMEPR in the population. The values reported in Fig.~\ref{fig:convergence_PMEPR_genes18} result from averaging out these minima over the multiple runs. This means that, for the case where $N$=100, after 400 generations in average one chromosome of the final population (one set of phase codes) generates an OFDM symbol whose PMEPR is 2.9. While Fig.~\ref{fig:convergence_PMEPR_genes18} shows the convergence of our GA for the case of M-PSK ($M$=$2^{18}$) Fig.\ref{fig:convergence_PMEPR_genes2} refers to the case of QPSK modulation. With Fig.\ref{fig:convergence_PMEPR_genes2}, we demonstrate that it is possible to control the PMEPR of an OFDM symbol that would rely on a predefined alphabet. In terms of the execution time, we found that in average, it took 115s to loop over 400 generations for the first case, when it took in average 104s for the second case, with $N$=100. When $N$=500, this time increases to 566s and 515s respectively. We are running our simulation on the university of Cape Town (UCT) server. Following, the preliminary analysis from Section~\ref{sec:sect2}, $N$=500 can be considered as an upper bound for pulsed OFDM radar. Hence, most likely, smaller number of subcarriers may be used, inferring computation times less than 500s. Whether the search can be implemented on the fly or not will depend on the infrastructure available in terms of the computation power. Otherwise, it may be done off line. 

\begin{figure}[!ht]
\centering
\mbox{\subfigure[M-PSK (18 genes)]{\includegraphics[width=1.8in]{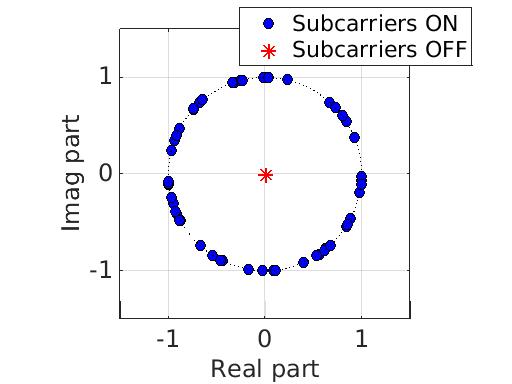}\label{fig:18genes_illustration_0_5_sparse}}
\subfigure[Sparse spectrum (50\%)]{\includegraphics[width=1.8in]{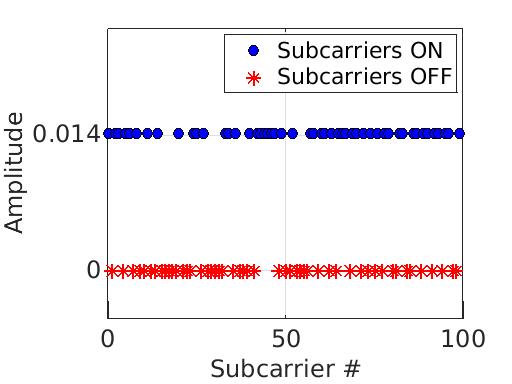}\label{fig:18genes_illustration_0_5_spectrum}}}
\mbox{\subfigure[QPSK (2 genes)]{\includegraphics[width=1.8in]{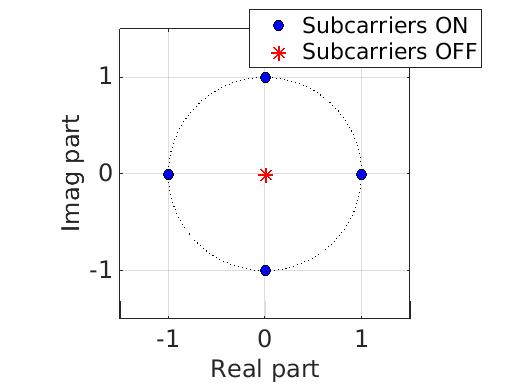}\label{fig:QPSK_illustration_0_7_sparse}}
\subfigure[Sparse spectrum (70\%)]{\includegraphics[width=1.8in]{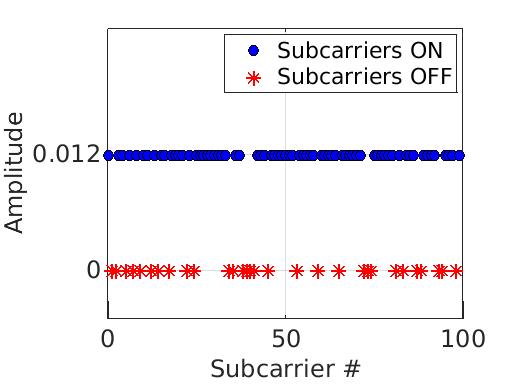}\label{fig:QPSK_illustration_0_7_sparse_spectrum}}}
\caption{Illustrating the complex phase codes and sparse spectrum in the case of, a) and b) M-PSK ($M$=$2^{18}$) and sparsity coefficient 50\% and in the case of, c) and d) QPSK and sparsity coefficient 70\%. We consider $N$=100 subcarriers.} \label{fig:Nomodulation_QPSK_illustration}
\end{figure}

Figs.~\ref{fig:18genes_illustration_0_5_sparse}-\ref{fig:QPSK_illustration_0_7_sparse} illustrate the complex phase codes corresponding to one chromosome of the final population for both cases. It is clear that a pseudo-continuum of values are available for the first case, while with QPSK, the arguments $\phi_{n,k}$ are limited to 4 values: 0, $\pi/2$, $\pi$ and $3\pi/2$.

The second limitation that we challenge with our GA based method relates to the case of sparse spectra. To that end, we repeated the above simulations for sparsely populated spectra, with sparsity coefficients of 70\% (out of 100 subcarriers, 70 subcarriers are ON, 30 are OFF) and 50\%. At each run of the Monte-Carlo simulation, a sparse spectrum is selected at random. In practice, we populate the band at random but maintain the two extreme subcarriers 'ON' to prevent from reducing 'badly' the bandwidth. Examples of spectra are given in Figs.~\ref{fig:18genes_illustration_0_5_spectrum}-\ref{fig:QPSK_illustration_0_7_sparse_spectrum} for the two sparsity options. Remark that, on the one hand, the weighting is uniform for all subcarriers 'ON' and on the other hand, its level is adjusted to provide an OFDM symbol of unit energy.  

 
For comparison purposes, we assessed the behaviour of the Newman phasing technique under the same conditions of sparse spectra. We also ran Monte-Carlo simulations and collected the mean value for the PMEPR over all runs. Taking the mean value makes sense since the decision as to which subcarriers shall be OFF may not be in the designer hands. We considered 1000~runs. Our results, given in linear unit, are presented in Table~\ref{table:PMEPR_results_N_100} for both cases $N$=100 and $N$=500. We observe that, under sparsity conditions, they outperform, in average, those obtained with Newman's technique. Note that, with $N$=500, we achieved better results in the constrained case of QPSK as opposed to the 'no~constellation' case, after 400 generations. One cause is the size of the search space, much bigger in the unconstrained case. Whether we can improve these results further would require optimizing our genetic algorithms for the case of large $N$. However, this falls outside the scope of this work.

\begin{table}[!ht]
\caption{Simulation results for the single objective PMEPR (in linear unit) 
with different levels of sparsity. Our GA based results are compared with two known results, viz., the trivial non-coded case and Newman phase codes' case} 
\centering 
\begin{tabular}{l | l | c | c | c | c } 
& & \multirow{2}{*}{Non-coded} & \multirow{2}{*}{Newman} & \multicolumn{2}{c}{GA} \\ [0.5ex]
& & & & $2^{18}$-PSK & QPSK \\ [0.5ex]
\cline{1-6}  
\multirow{3}{*}{\textbf{N=100}} & Full & 100 & 1.8 & 2.9 & 2.8 \\ 
\cline{2-6}
&70$\%$ & 70 & 4.3 & 3.1 & 3.1 \\
\cline{2-6}
&50$\%$ & 50 & 5.2 & 3.1 & 3.2 \\
\cline{1-6}
\multirow{3}{*}{\textbf{N=500}} & Full & 500 & 1.8 & 4.4 & 4.0 \\ 
\cline{2-6}
&70$\%$ & 350 & 5.0 & 4.4 & 4.1 \\
\cline{2-6}
&50$\%$ & 250 & 6.3 & 4.4 & 4.2 \\
\end{tabular}
\label{table:PMEPR_results_N_100} 
\end{table}

\begin{figure}[!ht]
\centering
\mbox{
\subfigure[Full band]{\includegraphics[width=1.8in]{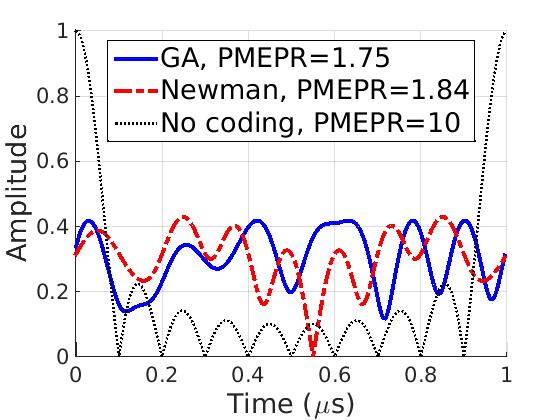}\label{fig:PMEPR_N_10_sparse_1}}
\subfigure[80$\%$ of the band]{\includegraphics[width=1.8in]{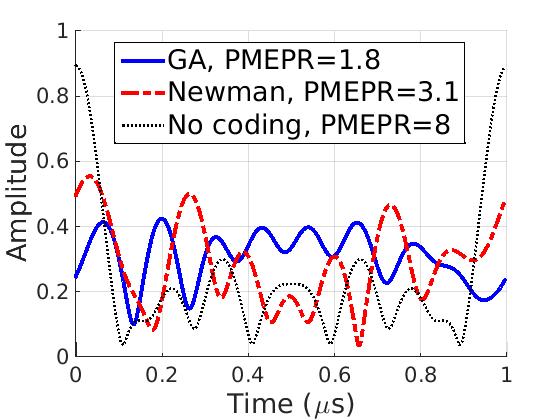}\label{fig:PMEPR_N_10_sparse_0_8}}}
\caption{Amplitude of the time domain OFDM symbol when a) all subcarriers are used and b) only 80$\%$ of the subcarriers are used, $N$=10. In both cases, our GA solution outperforms the Newman strategy. All pulses have unit energy. 
} \label{fig:PMEPR_N_10}
\end{figure}

We also show that our GA based technique can even outperform Newman's phasing method in the case of fully populated spectrum, with smaller number of subcarriers, viz., $N$=10 in this example. Fig.~\ref{fig:PMEPR_N_10_sparse_1} presents our PMEPR optimized OFDM symbol and compares it with the Newman based symbol. Fig.~\ref{fig:PMEPR_N_10_sparse_0_8} shows another PMEPR optimized OFDM symbol in the case of 80\% sparsity. The difference between both GA based and Newman based PMEPR is about 1~dB.

      
The third limitation raised by state-of-the-art techniques regards the lack of control over the spectrum, viz., the weights cannot follow any arbitrary pattern. We already observed this effect with the degradation of the results following Newman's technique in the case of sparse spectra. Sparsity is, to a certain extent, a case of non uniform weight pattern. Although the GA based results presented so far have been obtained with equal weights for all subcarriers ON, one could easily repeat the simulations including non uniform values. We propose to address separately this aspect in the context of a matched illumination procedure in Section~\ref{sec:sect6}. Our aim is to show that the PMEPR optimization does not impinge the matched illumination and reciprocally if they are addressed in the correct order.  

\subsection{Multiple objectives}\label{sec:sect5b} 

One benefit to use the NSGA-II in multi-objective optimization procedures lies in the fact that all objectives are optimized concurrently. The Pareto front created at each generation converges towards some optimum that provides the designer with the best set of solutions in a Pareto sense~\cite{Deb02}. In our OFDM pulse design problem, needs for a multi-objective optimization arise when the time domain processing would be selected as explained in Section~\ref{sec:sect3b}.

\subsubsection{PMEPR and PSLR}\label{sec:sect5b1}

\begin{figure}[!ht]
\centering
\mbox{\subfigure[$N$=25, $K$=4]{\includegraphics[width=1.8in]{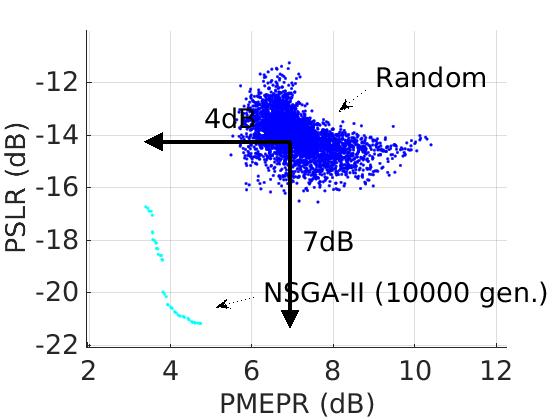}\label{fig:Comparison_after_1_and_10000_runs_N25_K4}}
\subfigure[$N$=125, $K$=4]{\includegraphics[width=1.8in]{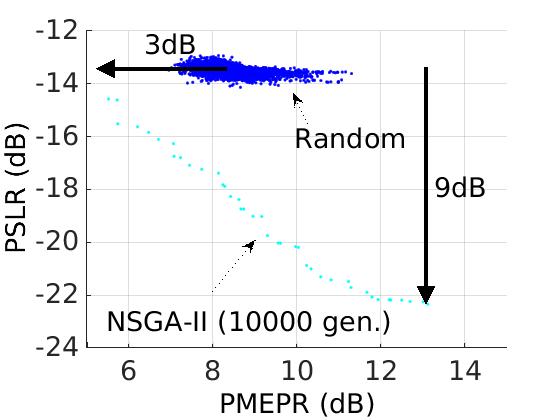}\label{fig:Comparison_after_1_and_10000_runs_N125_K4}}}
\mbox{\subfigure[$NK$=100]{\includegraphics[width=1.8in]{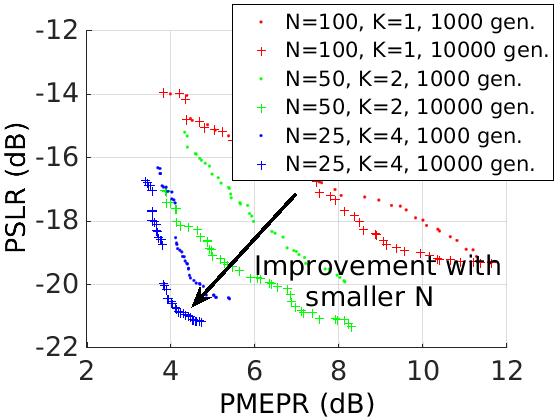}\label{fig:Comparison_after_1000_and_10000_runs_NK_100}}
\subfigure[$NK$=500]{\includegraphics[width=1.8in]{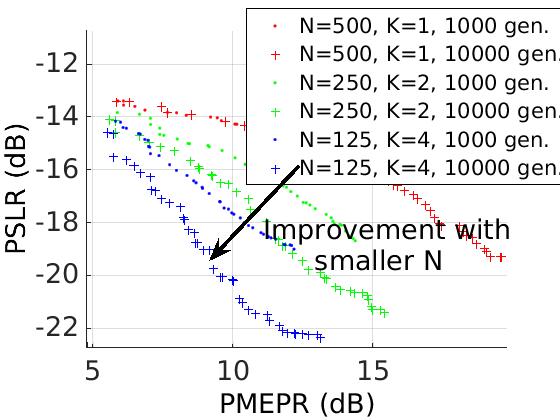}\label{fig:Comparison_after_1000_and_10000_runs_NK_500}}}
\caption{Comparison between random phases and phases resulting from the NSGA-II optimization in a) and b). Convergence of the NSGA-II optimization at constant time bandwidth products in c) and d).} \label{fig:time_ACF_pulses}
\end{figure}

We start by looking at the two-objective optimization with PMEPR and PSLR. In Figs.~\ref{fig:Comparison_after_1_and_10000_runs_N25_K4}-~\ref{fig:Comparison_after_1_and_10000_runs_N125_K4} we show how the NSGA-II can improve both objectives. We compare the Pareto front after 10000 generations to a random population.  
In both cases, we have taken a number of subcarriers smaller than 100 and 500, but the time bandwidth product remains the same, $NK$=100 and $NK$=500 respectively. 
The size of the population is equal to 40. 
After 10000 generations our set of optimal solutions is considerably improved. The Pareto front is visible and different sets of phase codes can be selected depending on whether the emphasis in on a low PMEPR or a low PSLR. In light of Fig.~\ref{fig:Comparison_after_1_and_10000_runs_N25_K4} we see that improvements as high as 7~dB and 4~dB can be achieved in terms of PSLR and PMEPR for the case of $N$=25, $K$=4. Likewise, Fig.~\ref{fig:Comparison_after_1_and_10000_runs_N125_K4} shows that improvements as high as 9~dB and 3~dB can be achieved in terms of PSLR and PMEPR for the case of $N$=125, $K$=4. The improvements reported here refer to the characteristics of the mean point in the random distributions.  

We then propose to assess the position of the Pareto front on the PMEPR-PSLR map for different number of subcarriers $N$ and compare three cases with equal time bandwidth products. We report the results in Figs.~\ref{fig:Comparison_after_1000_and_10000_runs_NK_100}-~\ref{fig:Comparison_after_1000_and_10000_runs_NK_500}. 
For each case, we plot the Pareto front after 1000 and 10000 generations. We note 
that designs with small $N$ will provide the best Pareto front. In these two-objective simulations, a cycle of 1000 generations takes about 70s and 370s for the cases, $N$=25 $K$=4 and $N$=125, $K$=4 respectively. 

\subsubsection{PMEPR constrained multi-objective optimization}\label{sec:sect5b2} 

\begin{figure}[!ht]
\centering
\mbox{\subfigure[PMEPR distribution (random)]{\includegraphics[width=1.8in]{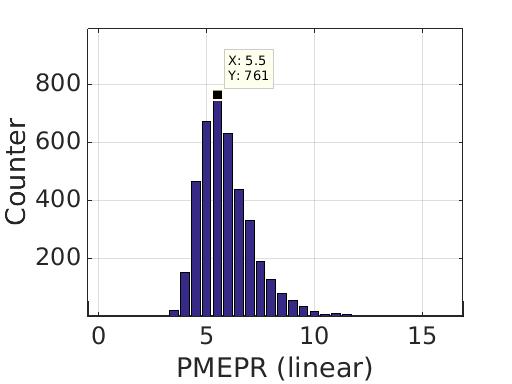}\label{fig:Hist_PMEPR_random_Montecarlo100_N100}}
\subfigure[No PMEPR constrain]{\includegraphics[width=1.8in]{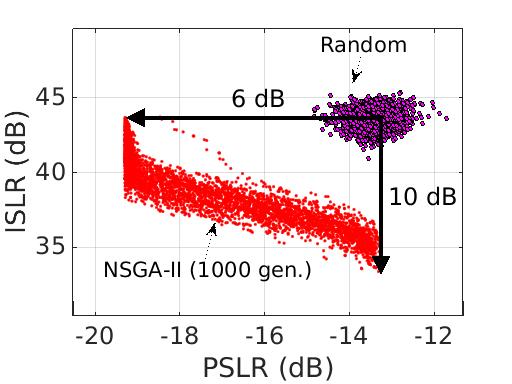}\label{fig:PSLR_ISLR_ParetoFronts1000generations_noconstrainPMEPR_Montecarlo100_N100}}}
\mbox{\subfigure[PMEPR distribution with constrain]{\includegraphics[width=1.8in]{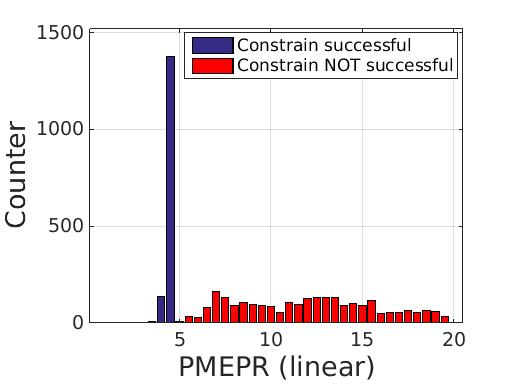}\label{fig:Hist_constrainedPMEPRmax5_Montecarlo100_N100}}
\subfigure[With PMEPR constrain]{\includegraphics[width=1.8in]{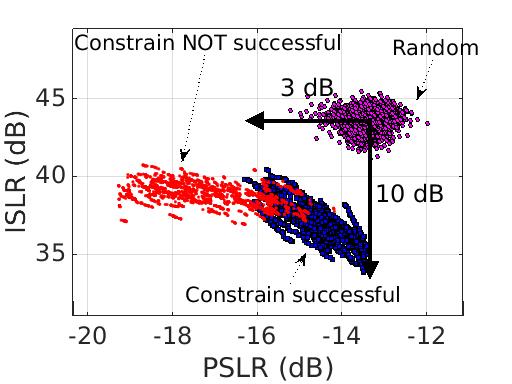}\label{fig:PSLR_ISLR_ParetoFronts1000generations_constrainedPMEPRmax5_Montecarlo100_N100}}}
\caption{a) Distribution of the PMEPR for random phase codes, $N$=100 b) 100 Pareto fronts obtained after 1000 generations with NSGA-II, no constrain on the PMEPR compared with random sets c) distribution of the PMEPR for all chromosomes from the last populations after 1000 generations with a constrain on the PMEPR d) 100 Pareto fronts obtained after 1000 generations with NSGA-II, with a constrain on the PMEPR equal to 5. The Pareto fronts where the constrain has been successful for all chromosomes of the population are in blue (38 out of 100).} \label{fig:}
\end{figure}

One may argue that it is more appropriate to treat the PMEPR as a constraint rather than an objective. Indeed, if the system provides a maximum value for the PMEPR, this may leave space for other objective functions, like the ISLR. As mentioned before, PSLR and ISLR pick up different behaviours of the sidelobe level whose impact will be scenario dependent. It is thus judicious to generate phase codes that can optimize both objectives in the form of Pareto fronts instead of treating each objective separately~\cite{Lellouch14a}. We report our preliminary scheme to deal with this constrained multi-objective problem as follows. Our idea has been to modify slightly the original version of the NSGA-II and in particular assign very low crowding distances to those chromosomes that result in PMEPR exceeding the threshold that we chose to apply. The motivation here is to reduce the chance for those chromosomes to participate to the evolution, such that over time, the population is composed of chromosomes with favourable PMEPR values. Although quite simple, this approach proved to work on the few cases analysed. Hereafter, we present the results of our analysis for the configuration $N$=100, $K$=1. In order to select a reasonable threshold value, we first assessed the PMEPR distribution, which we display in Fig.~\ref{fig:Hist_PMEPR_random_Montecarlo100_N100}. We selected the PMEPR value preceding the peak, 5 in this case. Our Monte-Carlo executed 100 runs of 1000 generations each and we analysed the characteristics of all 100 last populations. We considered the same parameter for the population size, 40. In Fig.~\ref{fig:Hist_constrainedPMEPRmax5_Montecarlo100_N100} we assess the PMEPR values for all elements of the last populations. We see that a big majority of chromosomes have their PMEPR below the threshold. Counting those populations that gave rise to 40 chromosomes with a PMEPR below the threshold, we get 38 cases out of 100. We can then plot the Pareto fronts after 1000 generations for all these 100 runs, distinguishing those fronts where the constrain is fully compliant and the other. We obtain the results as presented in Fig.~\ref{fig:PSLR_ISLR_ParetoFronts1000generations_constrainedPMEPRmax5_Montecarlo100_N100}. Comparing with the case where the optimization is run without constrain on the PMEPR, whose results are given in Fig.~\ref{fig:PSLR_ISLR_ParetoFronts1000generations_noconstrainPMEPR_Montecarlo100_N100}, we see that our constrained optimization gives Pareto fronts with reduced PSLR values but equally good ISLR values. 

\begin{figure}[!ht]
\centering
\mbox{\subfigure[No PMEPR constrain]{\includegraphics[width=1.8in]{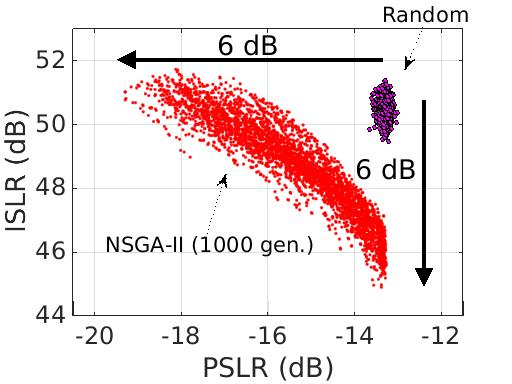}\label{fig:PSLR_ISLR_ParetoFronts1000generations_constrainedPMEPRmax6_5_Montecarlo100_N500}}
\subfigure[With PMEPR constrain]{\includegraphics[width=1.8in]{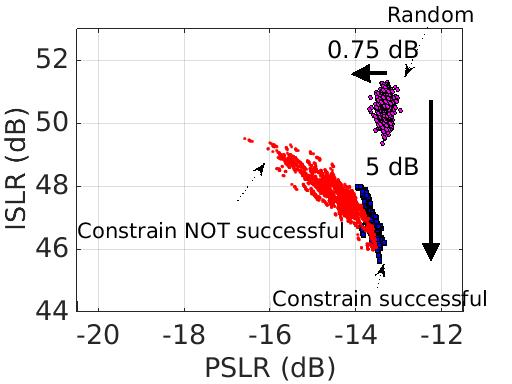}\label{fig:PSLR_ISLR_ParetoFronts1000generations_noconstrainPMEPR_Montecarlo100_N500}}}
\caption{a) 100 Pareto fronts obtained after 1000 generations with NSGA-II, no constrain on the PMEPR compared with random sets, $N$=500 b) 100 Pareto fronts obtained after 1000 generations with NSGA-II, with a constrain on the PMEPR equal to 6.5. The Pareto fronts where the constrain has been successful for all chromosomes of the population are in blue (22 out 100).} \label{fig:}
\end{figure}

For completeness, we repeated our analysis for the case $N$=500. The threshold for the PMEPR, selected in the same manner was chosen to be 6.5. Among the 100 final populations, only 22 fulfilled the criteria whereby all 40 chromosomes shall induce a PMEPR below 6.5. Comparing the results in the unconstrained case in Fig.~\ref{fig:PSLR_ISLR_ParetoFronts1000generations_noconstrainPMEPR_Montecarlo100_N500} with those from the constrained case as in Fig.~\ref{fig:PSLR_ISLR_ParetoFronts1000generations_constrainedPMEPRmax6_5_Montecarlo100_N500} we see that again, our constrained optimization gives Pareto fronts with similarly good ISLR values. The performance is not as great for the PSLR since the improvement is reduced from 6~dB down to 0.75~dB.

\section{Case study: GA for PMEPR optimization in a matched illumination procedure}\label{sec:sect6}
\begin{figure}[!ht]
\centering
\includegraphics[scale=0.35]{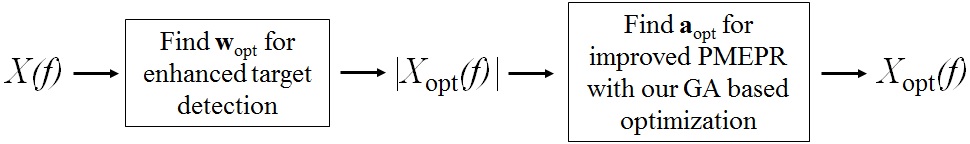} 
\caption{OFDM pulse optimization in two steps.} 
\label{fig:OFDM_opti_2steps}
\end{figure}

Matched illumination is a research direction that motivates OFDM radar~\cite{Genderen09}. In this section, our objective is to show how single objective GA based optimizations can be incorporated in a matched illumination procedure. 
We propose a two-step approach that addresses both problems of detection enhancement and PMEPR optimization sequentially. As such, this problem does not require having a multi-objective approach. The first step consists in finding an optimal weight vector $\mathbf{w_{\text{opt}}}$ to enhance the detection characteristics. In other words, it finds the power spectrum of the radar pulse. This vector is then passed on to our single objective GA based optimization to find a vector of phase codes $\mathbf{a_{\text{opt}}}$ (only one symbol is considered, hence we drop the index $k$) such that the resulting OFDM symbol has optimal PMEPR properties. 
Our procedure is presented in Fig.~\ref{fig:OFDM_opti_2steps}. The complex vector resulting from the element-wise multiplication $\mathbf{w_{\text{opt}}}\cdot\mathbf{a_{\text{opt}}}$ represents the discrete spectrum of our optimal OFDM pulse, which we define as $X_{\text{opt}}(f)$ in Fig.~\ref{fig:OFDM_opti_2steps}.

To understand how the weights of the OFDM signal can be used to enhance the target detection, we shall mention that in radar, scattering can be modelled as a linear filtering process where the received echo results from the convolution of the transmitted pulse with the target impulse response. Looking at it in the Fourier domain, we see that the spectrum of the received echo will follow from multiplying the spectrum of the transmitted pulse with the target reflectivity spectrum. Hence, the shape of the transmitted pulse spectrum, determined by the weights will impact the energy content of the received pulse. This last inference results from Parseval's theorem that links the energy of a time domain signal with the energy of its Fourier transform in the frequency domain.

\subsection{SNR as the design metric}\label{sec:sect6a}
Unlike the traditional approach to consider imaging based metrics for target detection and classification, Bell has introduced the signal-to-noise ratio (SNR) metric \cite{Bell93}. In that concept, the objective is to maximize the energy or SNR of the received echo for the particular target in the scene. The PSLR and ISLR are no longer relevant and can be ignored in the pulse design. In the following we revisit the formulation of this SNR maximization problem. Note that an important aspect concerns the definition of the matched filter.

\subsubsection{Transmitted signal receiver-filter pair}\label{sec:sect6b}
In \cite{Wilkinson98} the authors expressed the signal at the output of the matched filter $V(f)$ as the product of the target reflectivity spectrum $\varsigma(f)$, the pulse spectrum at baseband $X(f)$ and the filter transfer function $H(f)$:
\begin{equation} \label{eq:vBB}
V(f) =\varsigma(f+f_c)X(f)H(f).
\end{equation}

\begin{figure}[!ht]
\centering
\includegraphics[scale=0.4]{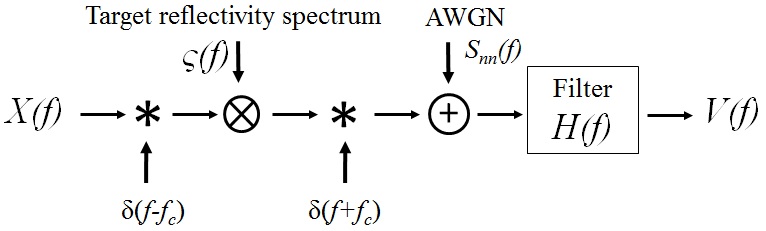} 
\caption{Model of the received signal in the frequency domain.} 
\label{fig:Signal_transformation_freq}
\end{figure}

In the frequency domain, the pulse goes through the steps as indicated in Fig.~\ref{fig:Signal_transformation_freq}. Both convolution symbols charaterize the up and down conversions. In the time domain, we can express the radio frequency (RF) analytical signal as $v_{tx}(t)$:

\begin{equation} \label{eq:vtx} 
v_{tx}(t) = x(t)\exp(j2\pi f_ct),
\end{equation}
where $f_c$ is the carrier frequency and $x(t)$ is the OFDM pulse as given in Eq.~\ref{eq:OFDMbaseband}. 

\subsubsection{Optimum receiver-filter pair}\label{sec:sect6b1}
In the case of an extended target, unlike the standard matched filtering problem applicable when the target is modelled as a point scatterer, the optimal receiver filter shall be matched to the waveform scattered by the target, not the transmitted target itself. When this is the case, the maximum SNR at time $t_0$ is given by the following equation \cite{Bell93}:
\begin{equation} \label{eq:SNR_opt} 
\left(\frac{S}{N}\right)_{t_0} = \int_{-\infty}^\infty \frac{|\varsigma(f+f_c)X(f)|^2}{S_{nn}(f)}df.
\end{equation}

While previous works have researched solutions to this SNR maximization problem in the time domain, in terms of $x(t)$, we propose to run the search in the frequency domain and look for $|X(f)|$ instead. This finally amounts to searching for an optimal weight vector $\mathbf{w_{\text{opt}}}$. If we assume that the spectrum of our OFDM pulse is zero outside the bandwidth\footnote{The spectrum of the baseband OFDM pulse is considered to spread from 0 to $B$ rather -$B$/2 to $B$/2}, and consider that the stationary additive Gaussian noise is white, with one-sided power spectral density $S_{nn}(f)=N_0$, we can modify Eq.~\ref{eq:SNR_opt} into: 
\begin{equation} \label{eq:SNR_opt_bb} 
\left(\frac{S}{N}\right)_{t_0} = \frac{1}{N_0}\int_0^B |X(f)\varsigma(f+f_c)|^2df,
\end{equation} 
Assuming further a deterministic expression for the target reflectivity spectrum, we can rewrite Eq.~\ref{eq:SNR_opt_bb} in discrete terms:
\begin{equation} \label{eq:SNR_opt_bb_discrete} 
\left(\frac{S}{N}\right)_{t_0} = \frac{1}{N_0}\sum_{n=0}^{N-1} |X[n]\varsigma[n]|^2,
\end{equation}  
where $X[n]$=$X(n\Delta f)$ and $\varsigma[n]$=$\varsigma(n\Delta f+f_c)$ are the baseband OFDM spectrum and the target reflectivity spectrum at RF, respectively, expressed in discrete terms. The index $n$ takes values from 0 to $N-1$. 
Thus, our optimization problem reduces in finding the weights $w_n$ according to:
  
\begin{equation} \label{eq:OFDMsymbolfreq}  
\text{arg }\underset{w_n}{\operatorname{max}} \sum_{n=0}^{N-1} w_n^2|\varsigma[n]|^2, \text{ s.t. } \left\{
    \begin{array}{ll}
        \mathbf{x}^T\mathbf{x}=1 \\
        v_{\text{l}} < w_n < v_{\text{u}}
    \end{array} 
    \right.
\end{equation}

The first constrain tells that the pulse energy remains unity throughout the search and the second constrain limits the dynamic range of the weight values. Different values for the lower and upper bounds will result in different spectrum characteristics. Although other methods may exist to solve this constrained single objective problem, we suggest applying our GA approach, here again. The expression with the summation is the new objective function and the variables are no longer the phase codes but the weights.   


\subsection{Simulation setup and results}\label{sec:sect6c}
In this analysis we invoke different dimensions for our OFDM pulse as compared to Section~\ref{sec:sect5}. We assume to work at X-band and consider a 2~GHz bandwidth $B$ centred around 10~GHz, $f_c$ = 9~GHz. Our OFDM pulse is composed of $N$=100 subcarriers and the subcarrier spacing is $\Delta f$=20~MHz. To appreciate the detection enhancement resulting from the optimal weight vector, we must normalize the target reflectivity spectrum $\varsigma[n]$. 

\subsubsection{Target reflectivity spectrum normalization}\label{sec:sect6c1}
We normalize the discrete target reflectivity spectrum following the strategy presented in~\cite[chap.~14]{Gini12}. The later states that when a flat spectrum OFDM pulse of unit energy interacts with the normalized reflectivity spectrum $\varsigma_{\text{norm}}[n]$, the frequency-domain reflected signal shall have unit average power, viz., $1/N\cdot\sum_{n=0}^{N-1} w_n^2|\varsigma_{\text{norm}}[n]|^2$=1. The weights corresponding to our flat spectrum OFDM pulse of unit energy are given\footnote{In fact, different values are possible depending on how one implements the IDFT algorithm, and in particular whether a multiplicative constant is applied} by $w_n=1/\sqrt{N}$. Hence, we find:
\begin{equation} \label{eq:H_norm} 
\varsigma_{\text{norm}}[n] = \frac{N}{\sqrt{\sum_{n=0}^{N-1} |\varsigma[n]|^2}}\cdot \varsigma[n],
\end{equation}  
 
\subsubsection{Complex target model}\label{sec:sect6c2}
We simulate a synthetic target composed of $P$=50 point scatterers, each with the same reflectivity $\varsigma_i$=$\sqrt{\sigma_i}$=1 and located at ranges $R_i$ from the radar. The individual point scatterers are assumed to be perfectly conducting spheres, large enough to have a reflectivity constant within the band of interest. It can be shown \cite{Richards10} that the compound target reflectivity spectrum $\varsigma(f)$ is equal to: 
\begin{equation} \label{eq:reflectivity_complex_freq} 
\varsigma(f) = \sum_{i=1}^P\sqrt{\sigma_i}\exp(-j4\pi f\frac{R_i}{c}).
\end{equation}
The point scatterers are randomly distributed within a rectangle, 5~meters wide and 10~meters long, whose center is 10~km away from the radar along the x axis. 


\subsubsection{SNR and PMEPR improvements}\label{sec:sect6c3}
\begin{figure}[!ht]
\centering
\mbox{\subfigure[Optimal weight vector]{\includegraphics[width=1.8in]{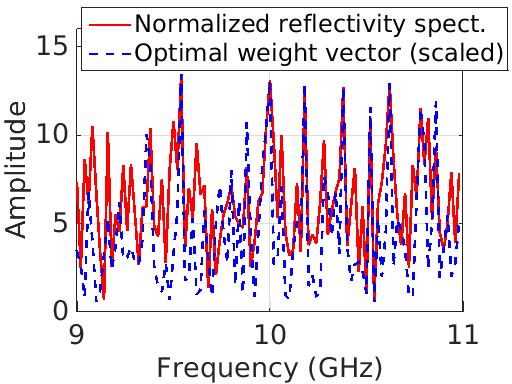}\label{fig:Target_Reflectivity_OptWeights}}
\subfigure[PMEPR improvement]{\includegraphics[width=1.8in]{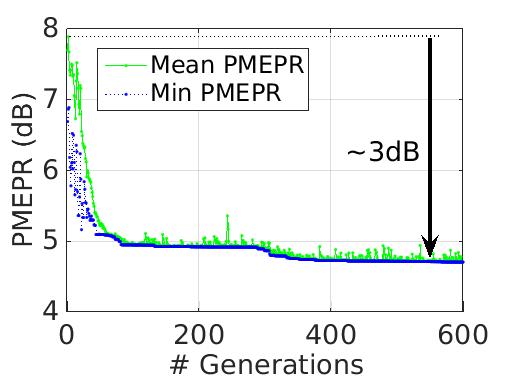}\label{fig:PMEPR_convergence_v2}}}
\caption{a) Optimal weight vector $\mathbf{w}_{\text{opt}}$ versus the normalized reflectivity spectrum resulting from a complex target made of $P$= 50 point scatterers and b) convergence of the single GA based technique to find an optimal set of phase codes $\mathbf{a}_{\text{opt}}$.} \label{fig:twostep_optimization}
\end{figure}

Following the above methodology, we first generate an optimal weight vector $\mathbf{w_{\text{opt}}}$. We employ a continuous GA~\cite{Haupt04} with 20~chromosomes, a mutation rate of 0.2 and select arbitrarily $v_{\text{l}}$=0.01 and $v_{\text{u}}$=10. We also include in the initial population a scaled version of the vector $\varsigma_{\text{norm}}$, since we know that this is a good solution. We stop the GA after 5000~generations, and consider the final weight vector to be $\mathbf{w_{\text{opt}}}$. This takes about 50s. Fig.~\ref{fig:Target_Reflectivity_OptWeights} puts on the same graph the normalized reflectivity spectrum for our complex target as well as the optimal weight vector, scaled to fall within the same boundaries. We see that most of the energy is distributed on those subcarriers that correspond to frequencies where the reflectivity spectrum is strong. 
We present the gain resulting from 
this optimal weight vector 
in table~\ref{tab:result_enhancement}. The gain is a measure of the difference in average power with the case of a flat OFDM spectrum and is equal to 2.8~dB. This value results from averaging over 10 runs in a Monte-Carlo simulation. 

Now that we have fixed $|X_{\text{opt}}|$ the second step consists in feeding $\mathbf{w_{\text{opt}}}$ into our second GA where the objective function is the PMEPR. We consider again 12~chromosomes in the population and run it for 600~generations. This takes about 170s. A measure of the convergence of our solution is presented in Fig.~\ref{fig:PMEPR_convergence_v2}. We find the best element of the final population and define it to be $\mathbf{a_{\text{opt}}}$. 
We observe that the PMEPR has been improved by 3~dB. This results is in agreement with our results from Section~\ref{sec:sect5a}. In the end, as shown in Fig~\ref{fig:OFDM_opti_2steps} our pulse $X_{\text{opt}}(f)$ has enhanced detection capabilities for the target of interest and optimal PMEPR. 

\begin{table}[h]
\caption{Result for the detection enhancement. The gain is measured in terms of average power} 
\centering 
\begin{tabular}{c | c | c} 
  & Flat spectrum & GA based $\mathbf{w}_{\text{opt}}$
\\ [0.5ex]
\hline 
Gain (dB)& 0 & 2.8 \\ 
\hline 
\end{tabular}
\label{tab:result_enhancement}
\end{table}
 
\section{Conclusion}\label{sec:sect7}

In this paper, we showed that GA based techniques are suitable to optimize or improve several features of the OFDM pulses, provided reasonable search space, hence, reasonable numbers of subcarriers (and symbols). In pulsed OFDM radar though, we expect limited numbers of subcarriers as a result of the orthogonality principle that links the symbol duration to the subcarrier spacing. We inspected the possibility to incorporate these optimization techniques in a waveform design framework in regard of two processing solutions, specific to OFDM radar.  

When dealing with the frequency domain processing solution, we pointed out that the focus shall be the minimization of the PMEPR. 
This figure-of-merit is a salient aspect of the OFDM pulse whose time domain signal is subject to strong variations. In this context, we suggested using a single objective GA. We showed that this evolutionary technique can produce solutions that outperform robust state-of-the-art methods. In particular, we challenged several limitations of those methods and demonstrated the merit of our approach for the case 1) when the spectrum is composed of non contiguous subcarriers and 2) when the phase codes are constrained to discrete values, as it is the case with specific constellations.

Next, we explained that the time domain processing solution requires optimal autocorrelation sidelobe and PMEPR properties. 
We showed that it is possible to apply another evolutionary algorithm, namely the NSGA-II to search for optimal sets of phase codes in a Pareto sense. We observed that substantial improvement is achieved in comparison to random coding. The Pareto optimal solutions obtained with this technique allow various options for different design preferences. We observed that, at a fixed time bandwidth product, the case with the smallest number of subcarriers produces the best Pareto front. Next, we looked at the possibility to treat the PMEPR as a constrain instead, while the two objectives are the PSLR and the ISLR. We demonstrated that our modified NSGA-II could produce Pareto fronts with equivalent characteristics in terms of the ISLR as for the unconstrained case. The same does not hold true for the PSLR, since the improvement becomes less in the constrained case.

Ultimately, we presented a case study where we incorporated the PMEPR optimization technique into a matched illumination strategy. 
We showed that both optimization problems could be decoupled, if the weights are reserved for the detection enhancement part while the phase codes are saved for the PMEPR part. We demonstrated that the enhanced detection problem is equivalent to finding an optimal weight vector and indicated that here again, a single GA could be used. We characterized the enhanced detection in terms of the metric, namely, average power, that gives 0~dB when the OFDM pulse has a flat spectrum of unit energy. Then, we illustrated the convergence of the PMEPR and found that we could obtain optimal sets of phase codes to improve the PMEPR by about 3~dB in comparison to the case where random codes would be used.

Overall, our optimization methods run from few 10s of seconds to few 100s of seconds, hence real time operations would require platforms with high processing power. Alternatively, they can be run offline. The results are then stored in lookup tables for future use. Improvement of the execution time for our GA based techniques and optimization of the convergence are directions for future work. Alternative optimization strategies, among other, convex methods are also part of the next research directions.  

\bibliographystyle{IEEEtran}
\bibliography{IEEEabrv,paper}

\end{document}